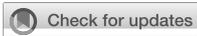

# ORCH: many analyses, one merge—a deterministic multi-agent orchestrator for discrete-choice reasoning with EMA-guided routing

Hanlin Zhou[1,2] and Huah Yong Chan[1]*

[1]School of Computer Sciences, Universiti Sains Malaysia, Gelugor, Malaysia, [2]Xiamen Institute of Software Technology, Xiamen, China

**Introduction:** Multi-agent/ensemble approaches can improve discrete-choice reasoning with large language models, but common orchestration methods are often non-deterministic, expensive, and difficult to reproduce. We propose ORCH, a deterministic multi-agent orchestrator that targets higher accuracy and better cost–performance via stable routing.

**Methods:** ORCH uses a pool of heterogeneous LLM agents and a deterministic routing mechanism based on exponential moving average (EMA) performance tracking. For each question, ORCH selects a small subset of agents, obtains candidate answers, and merges them through a controlled aggregation procedure. We evaluate ORCH on multiple discrete-choice benchmarks and compare against single-model baselines and non-routed ensemble strategies under consistent prompting and scoring.

**Results:** ORCH delivers consistent accuracy improvements over the best low-cost single model and provides additional gains over high-cost single-model baselines on several tasks, while reducing reliance on always-invoking expensive models. The deterministic routing and merge pipeline improves stability across runs.

**Discussion:** ORCH demonstrates that deterministic EMA-guided routing can offer a practical and reproducible orchestration strategy for discrete-choice reasoning. This framework can be extended to additional tasks, agent pools, and preference-aware routing policies in future work.

# 1 Introduction

Artificial intelligence is commonly characterized as an entity capable of perceiving its environment and context, and then making appropriate decisions on that basis. Within this landscape, agents are widely regarded as a core building block for constructing artificial general intelligence (AGI). The emergence of large language models (LLMs) has substantially expanded the functional space and development potential of such agents, particularly in information access, interaction, and complex reasoning (Baroni et al., 2022). Early widely known LLMs include the ChatGPT family (from ChatGPT-3 to ChatGPT-4), followed by other models built on similar natural language processing (NLP) techniques such as DeepSeek, Claude, and Gemini; most of these systems are designed as general-purpose models that can support a broad range of tasks across multiple domains. As model size has grown, researchers have identified so-called scaling laws, which show that performance tends to improve in a predictable manner as the number of parameters, the volume of training data, and the amount of compute increase (Xia Y. C. et al., 2024; Pezeshkpour et al., 2024). Contemporary commercial models such as GPT-4 not only provide strong language understanding and generation capabilities, but also exhibit emergent behaviors, including in-context learning, few-shot reasoning, and the ability to decompose complex tasks, thereby laying a technical foundation for the development of advanced agent systems (Pezeshkpour et al., 2024).

Built on top of modern LLMs, agent-based systems are no longer mere question–answer tools; they are increasingly designed as entities capable of autonomous planning, tool use, and multi-step reasoning. An agent can decompose a complex user goal into a sequence of executable sub-tasks, invoke external tools and API on demand, and maintain long-horizon contextual information through explicit memory mechanisms (Gao et al., 2024). For instance, the Chain-of-Agents framework employs multi-agent distillation and reinforcement learning to teach models when to call tools and how to schedule intermediate reasoning steps. In combination, these capabilities enable agents to provide practical support in domains such as software development, scientific assistance, and decision making (Fasterholdt et al., 2022). But, a single agent still exhibits clear limitations on challenging tasks, including bounded capabilities, restricted context windows, incomplete knowledge coverage, and limited robustness. These issues become particularly acute in problems that require expertise from multiple domains and careful coordination over many steps, where a single agent often fails to reach a satisfactory performance level.

To address these shortcomings, multi-agent systems (MAS) have emerged as a natural extension. By coordinating multiple specialized agents, such systems can perform task decomposition, parallel execution, and complementary optimization, thereby substantially improving overall performance and robustness. A number of representative frameworks have attracted attention in both research and industrial settings. MetaGPT (Hong et al., 2023), for example, encodes software-engineering workflows as standard operating procedures (SOPs) and assigns distinct roles—such as product manager, architect, and engineer—to different agents, effectively automating large parts of the development pipeline. Modern multi-agent systems typically rely on several core mechanisms, including dynamic agent creation (instantiating specialized agents on demand according to task requirements), shared memory management (allowing agents to read from and update a common knowledge repository), and routing policies that determine how tasks are assigned and passed between agents. For example, RCR-Router employs a role-aware context routing strategy that reduces token consumption by about 30% while maintaining output quality (Talebirad and Nadiri, 2023). Empirical studies further indicate that multi-agent architectures can deliver both performance and efficiency gains on various benchmarks; in particular, parallel multi-agent designs have been shown to achieve more than a twofold speed-up while improving resource utilization (Thiyagalingam et al., 2022). Despite these advantages, multi-agent systems still face several challenges, such as the complexity of integrating heterogeneous agents, coordination overheads arising from inter-agent communication and synchronization, increased computational cost, and the lack of unified evaluation protocols and interoperability standards.

Most existing multi-agent frameworks determine agent selection and task allocation using heuristic rules or stochastic routing strategies. While such approaches can be effective in some settings, they suffer from clear limitations: the inherent randomness makes system behavior difficult to reproduce, the lack of transparency hampers debugging and optimization, and in high-stakes domains such as medical diagnosis or financial decision making, uncontrolled stochasticity may introduce unacceptable risk. These issues are particularly acute for discrete-choice reasoning tasks, where the system must select a single option from a finite set of candidates; in this context, routing randomness and decision uncertainty can substantially degrade decision quality and erode user trust.

To address these concerns, this work proposes a rule-level determinism multi-agent orchestrator specifically designed for discrete-choice reasoning. The core idea is to adopt a "decompose-then-aggregate" paradigm, where the original problem is first split into analyzable components and the final answer is then produced by a dedicated merge process. In addition, to further improve performance, we optionally incorporate an exponential moving average (EMA)–guided routing strategy. This mechanism leverages historical performance statistics to adaptively choose among candidate agents, thereby optimizing agent selection in an online manner while preserving determinism at the system level.

The main contributions of this work can be summarized as follows.

(1) We design a protocol-level deterministic multi-agent orchestration framework for discrete-choice reasoning, which replaces heuristic or stochastic routing with fixed task decomposition, agent assignment, and answer aggregation policies, thereby reducing randomness in agent behavior and improving the reliability, reproducibility, and interpretability of the overall system.
(2) We optionally introduce an EMA-guided routing module that updates agent selection using historical accuracy/latency/cost; since it requires answer-based feedback, it is mainly intended for benchmarking, controlled evaluation, or delayed-feedback settings, and live deployment would require proxy feedback signals in place of gold labels.
(3) We conduct comprehensive experiments on multiple benchmark datasets to evaluate both effectiveness and efficiency, and provide systematic comparisons against strong single-model baselines and existing multi-agent or ensemble methods, together with ablation studies that quantify the contribution of each component.

The remainder of this article is organized as follows. Section 2 reviews related work on multi-agent systems and discrete-choice reasoning. Section 3 presents the design principles of the proposed protocol-level deterministic orchestration framework and details the EMA-guided routing mechanism. Section 4 describes the experimental setup, benchmark datasets, and empirical results. Section 5 discusses the limitations of the framework and outlines possible directions for future improvements, and the paper and highlights potential avenues for real-world deployment and applications.

# 2 Literature review

## 2.1 The development of large language models

Artificial Intelligence (AI) as a discipline has undergone multiple developmental stages since its formal establishment at the Dartmouth conference in 1956 (Baroni et al., 2022). Early AI research was primarily focused on symbolism and knowledge representation techniques, typified by work on expert systems and rule-based reasoning systems. During this period, AI systems sought to simulate human expert decision-making in specific domains by using explicitly encoded rules and a knowledge base. Although these systems achieved some success, they encountered significant challenges—notably the "knowledge acquisition bottleneck" and insufficient generalisation abilities.

From the late 1980s through the 1990s, machine learning (Machine Learning, ML) gradually became the mainstream paradigm in AI research. By automatically extracting patterns and regularities from data, ML approaches alleviated some of the limitations of traditional symbolic methods. Classical algorithms such as decision trees, support vector machines, and Bayesian networks were widely investigated and applied during this period. Entering the twenty-first century, the rapid growth of computational power together with the availability of large-scale datasets triggered a transformative shift driven by deep learning (Deep Learning) (Choudhury et al., 2024). This agentic perspective is increasingly regarded as a promising technical pathway toward artificial general intelligence (AGI), as it enables AI systems to integrate perception, reasoning, decision-making, and long-horizon control within a unified framework (Xia Y. et al., 2024).

The introduction of BERT in 2018 marked the maturation of the pre-training–fine-tuning paradigm in natural language processing (Ramos et al., 2022). By being pre-trained on large volumes of unlabelled text, BERT acquires rich contextual language representations, which in turn lead to substantial improvements on a wide range of natural language understanding tasks. Subsequently, the development of the GPT series further demonstrated the strong potential of generative pre-training (Nathalia et al., 2023). From GPT-1 to GPT-3, model size increased from 117 million to 175 billion parameters, empirically illustrating the effect of so-called scaling laws: model performance improves in a predictable manner as the number of parameters, the amount of training data, and the available computational resources grow (Amirhossein, 2023).

The release of GPT-3 drew significant attention to the phenomenon of emergent capabilities (Floridi and Chiriatti, 2020). Studies have shown that once a model surpasses certain scale thresholds, qualitatively new abilities arise that are absent in smaller models, such as in-context learning and few-shot reasoning. These capabilities enable large language models to accomplish novel tasks without task-specific training, relying only on a small number of examples, thereby indicating concrete progress toward more general forms of intelligence (Lyu et al., 2019).

More recent models, including GPT-4, Claude, and Gemini, further extend the capability frontier of large language models (Ouyang et al., 2022). Beyond strong performance in language understanding and generation, these models exhibit advanced skills in multi-modal comprehension, complex reasoning, and code generation. In tasks that demand multi-step reasoning and integration of dispersed knowledge, they can approach human-level performance. Nevertheless, despite the rapid gains in the capabilities of individual large language models, they still face persistent challenges such as hallucination, delayed or static knowledge updates, and limited consistency in reasoning (Xie et al., 2024).

## 2.2 Development of multi-agent systems

Multi-agent systems address complex problems that are difficult for a single agent to solve by coordinating multiple specialized agents (Junda et al., 2025). The core advantages of this paradigm include: (1) specialized division of labour, where different agents can focus on particular subtasks; (2) parallel processing, as multiple agents can operate simultaneously to improve efficiency; (3) enhanced robustness, since redundancy and complementary capabilities reduce the risk of single points of failure; and (4) scalability, enabling the number of agents to be dynamically adjusted according to task requirements (Talebirad and Nadiri, 2023).

MetaGPT is a representative example of multi-agent collaboration, in which the software development process is modeled as cooperation among multiple roles (Hong et al., 2023).

ChatDev adopts a similar multi-role collaboration paradigm but places greater emphasis on iterative development and quality assurance. Through a cyclical process of design, coding, testing, and review, different agents contribute their expertise at each stage (Qian et al., 2024).

AutoGPT and AgentGPT explore more autonomous forms of multi-agent systems (Yang et al., 2023). These systems can decompose high-level objectives into sub-tasks, assign them to specialized agents, monitor execution progress, and adapt plans dynamically.

MegaAgent demonstrates a large-scale collaborative system comprising hundreds of agents (Wang Q. et al., 2024). By introducing a hierarchical management architecture and dynamic task allocation mechanisms, the system can effectively coordinate a large number of agents, avoiding the communication bottlenecks and coordination chaos that arise in traditional settings. Experimental evidence suggests that, under appropriate organizational structures, increasing the number of agents can indeed lead to performance gains, thereby supporting the "collaboration at scale" hypothesis.

AgentVerse offers a general-purpose framework for multi-agent collaboration that supports multiple coordination patterns and communication mechanisms (Chen W. et al., 2023). Its modular design allows researchers to conveniently experiment with diverse collaboration strategies, thereby facilitating further research and practical applications of multi-agent systems as Table 1.

## 2.3 Agent routing and orchestration strategies

In multi-agent systems, selecting an appropriate agent for a given task is a core problem, and routing strategies strongly affect performance, efficiency, and reliability (Canese et al., 2021). Existing work mainly considers the following approaches.

Random routing selects one or more agents uniformly at random (Ong et al., 2024). It is easy to implement but ignores agent specialization and leads to non-reproducible results.

TABLE 1 Comparison of representative multi-agent frameworks.

| Framework | Year | Core mechanism | Application domain | Limitation |
| --- | --- | --- | --- | --- |
| MetaGPT | 2023 | Standardized operating procedures (SOPs) | Software development | Fixed workflow, limited flexibility |
| ChatDev | 2023 | Iterative development cycle | Software engineering | High communication overhead |
| AutoGPT | 2023 | Autonomous goal decomposition | General task execution | Limited controllability |
| MegaAgent | 2024 | Hierarchical management architecture | Large-scale collaboration | High architectural complexity |
| AgentVerse | 2024 | Modular multi-agent framework | General multi-agent platform | Requires expert configuration |

Input label scheme.

Rule-based routing uses hand-crafted rules to match tasks with agents, for example by task type, keywords, or domain labels (Rubak and Taheri, 2023). It is interpretable but requires substantial expert effort and adapts poorly to new task types.

Performance-based adaptive routing chooses agents according to their past performance (William et al., 2024). Multi-Armed Bandit algorithms are often used to balance exploration and exploitation, enabling online adaptation, but convergence can be slow.

Context-aware routing incorporates task content and context into routing decisions (Liu et al., 2025) For example, RCR-Router uses role-aware, context-sensitive routing based on dialogue history and task requirements, while reducing token usage by about 30% without harming accuracy.

Overall, existing routing strategies share several limitations: many are non-deterministic and thus hard to reproduce, deep learning–based routers are often opaque, learning-based methods incur high training costs, and fixed or static policies struggle to adapt to changing task distributions.

The main challenges can be summarized as follows:

(1) Integration complexity: the absence of unified interface standards among heterogeneous agents makes system integration difficult;
(2) Coordination overhead: communication costs, synchronization delays, and conflict-resolution mechanisms become particularly significant in large-scale systems;
(3) Scalability: as the number of agents grows, communication bottlenecks and topology-dependent issues become increasingly prominent;
(4) Insufficient determinism: many existing systems rely on heuristic rules or random routing strategies, leading to non-deterministic behavior and poor reproducibility of results;
(5) Lack of evaluation standards: the absence of unified performance metrics and interoperability specifications makes it difficult to compare different systems in a consistent manner.

Table 2 summarizes existing routing and aggregation styles; building on these observations, we introduce ORCH as a protocol-level deterministic orchestrator that resolves multi-analysis conflicts with a budget- and latency-aware control loop.

## 2.4 Discrete-choice reasoning and evaluation benchmarks

Discrete-choice reasoning refers to tasks in which a model must select the correct answer from a finite set of predefined options. This constitutes an important paradigm for assessing the reasoning capabilities of language models. Compared with open-ended generation, discrete-choice tasks offer clear evaluation criteria and make it easier to quantitatively compare the performance of different models and methods.

Massive Multitask Language Understanding (MMLU) is one of the most widely used multi-domain evaluation benchmarks (Zhao et al., 2025) It comprises 15,908 four-option multiple-choice questions spanning 57 subject areas, including STEM, the humanities, and the social sciences.

MMLU-Pro is an enhanced version of MMLU that increases the number of answer options (from four to ten) and raises the difficulty of the questions to provide a more challenging benchmark (Wang Y. et al., 2024).

Grade School Math 8K (GSM8K) focuses on mathematical reasoning (Zhong et al., 2026). GSM8K emphasizes multi-step numerical calculation and logical reasoning, and has become a key benchmark for evaluating chain-of-thought style reasoning in language models. Despite its seemingly elementary scope, many large language models still have considerable room for improvement on this benchmark.

The AI2 Reasoning Challenge (ARC) comprises science questions and is divided into Easy and Challenge subsets (Bhakthavatsalam et al., 2021). ARC-Challenge specifically targets questions that require more complex reasoning and substantial background knowledge, thus imposing higher demands on a model scientific reasoning capabilities.

HellaSwag is designed to evaluate commonsense reasoning and situational understanding (Li et al., 2025). The task asks models to choose the most plausible continuation of a short story from four candidates, requiring an understanding of everyday scenarios and causal relationships.

Taken together, these benchmark datasets form a standard test suite for evaluating discrete-choice reasoning. They differ in difficulty, domain coverage, and reasoning focus, thereby providing a multi-dimensional perspective for the comprehensive assessment of models and methods.

## 2.5 Research gap

Most existing multi-agent system (MAS) frameworks rely on heuristic rules or stochastic routing to select agents and assign tasks, which makes system behavior highly variable and leads to unstable accuracy—an unacceptable property in safety-critical domains such as medical diagnosis and financial decision-making. In addition, current agent selection mechanisms are not designed in a systematic,

TABLE 2 Comparison of agent routing strategies.

| Strategy type | Adaptive | Computational cost | Interpretability | Representative work |
|---|---|---|---|---|
| Random routing | No | Low | Low | Random selection |
| Rule-based routing | Yes | Low | High | Rule-based matching |
| Learning-based routing | Yes | High | Low | Routing transformer |
| Performance-adaptive routing | No | Medium | Medium | Multi-armed bandit |
| Context-aware routing | Yes | Medium | Medium | RCR-router |
| EMA-guided routing (this work) | Yes | Low | High | ORCH + EMA |

Dataset sources and basic information.

data-driven manner: agents are typically chosen via static configurations or simple hand-crafted rules, without making full use of the rich historical performance signals that could be used for dynamic optimization, resulting in wasted computational resources and suboptimal overall performance. Invoking all available agents by default incurs substantial computational cost and latency, whereas relying on a single agent cannot guarantee decision quality, and existing systems rarely provide mechanisms to adjust the level of concurrency according to task characteristics and agent confidence. Finally, there is a lack of specialized support for discrete-choice reasoning. Although many MAS frameworks perform well on general-purpose tasks, relatively few are explicitly tailored to multiple-choice questions or other finite-option decision problems, and dedicated collaboration strategies and evaluation protocols for such scenarios remain limited. In response to these limitations, this study introduces a protocol-level deterministic multi-agent orchestrator that combines EMA-guided adaptive routing, risk-aware scoring, and dynamic concurrency control, thereby providing a structured solution to these open challenges.

## 3 Methodology

Based on the research gaps identified in the literature review, this study proceeds along four main lines. First, it designs a multi-agent architecture together with the corresponding orchestration algorithms, enabling several AI agents to coordinate effectively while being evaluated on carefully selected benchmark datasets for accuracy assessment. Second, informed by the initial experimental results, an exponential moving average (EMA)–based routing mechanism is incorporated and re-evaluated. Third, self-consistency–enhanced variants are further examined to analyse their impact on discrete-choice reasoning. Finally, the study conducts a systematic comparison between the full multi-agent framework and a single–AI-agent baseline, in order to isolate the contribution of each functional module as Figure 1.

Determinism in this paper refers to the rule-/protocol-level coordination logic of ORCH: under a fixed evaluation protocol and fixed agent roles, ORCH uses fixed routing and aggregation rules (with deterministic tie-breaking) and, unless stated otherwise, deterministic decoding EMA-guided routing is a heuristic layer whose selections may change as historical feedback accumulates, although its update/selection rule is reproducible given a fixed protocol and fixed history. Self-consistency is an optional stochastic variant $K = 2$ that intentionally relaxes strict determinism and is reported separately (e.g., ORCH + SC).

### 3.1 Research question and overall framework

This study investigates the following question: without retraining the underlying LLM, can a multi-agent orchestration scheme consistently improve performance on multiple-choice reasoning and mathematical reasoning tasks, while simultaneously controlling inference latency and invocation cost? To this end, we design a protocol-level deterministic multi-agent orchestration, denoted ORCH, which follows a "Many Analyses, One Merge" paradigm: several agents independently analyse the same problem, and a dedicated merger integrates their outputs into a single discrete-choice decision. The overall framework is illustrated in Figure 2.

We compare heuristic routing with stochastic routing using worked examples, and we clarify how ORCH differs. Routing strategies can be grouped into two common families. Heuristic routing uses explicit, human-specified decision rules that map observed signals to an action; typical examples include rule tables (e.g., "if domain = math then route to agent A"), threshold gates (e.g., "if estimated difficulty > $\tau$ then allocate a stronger agent"), or deterministic fallbacks (e.g., "if verification fails then invoke a checker agent"). In contrast, stochastic routing explicitly introduces randomness into the selection step; examples include sampling an agent according to a probability distribution (softmax sampling), $\varepsilon$-greedy selection (usually pick the current best agent, but explore a random alternative with probability $\varepsilon$), or randomized bandit-style choices where exploration is driven by uncertainty rather than fixed thresholds.

How ORCH differs from heuristic/stochastic routing. ORCH implements a reproducible discrete-choice routing rule (e.g., argmax with a fixed priority/tie-break) under protocol-level deterministic decoding (temperature = 0). EMA, when enabled, is used only to smooth routing signals over time (stabilizing the decision boundary) rather than to perform probabilistic sampling; it therefore does not convert ORCH into a stochastic router. Empirically, adding EMA yields a small numerical improvement in confusion-matrix result, but the paired comparisons do not reach statistical significance on the evaluated MMLU setting; accordingly, we report the effect as a modest trend on our chosen configurations rather than a universal guarantee.

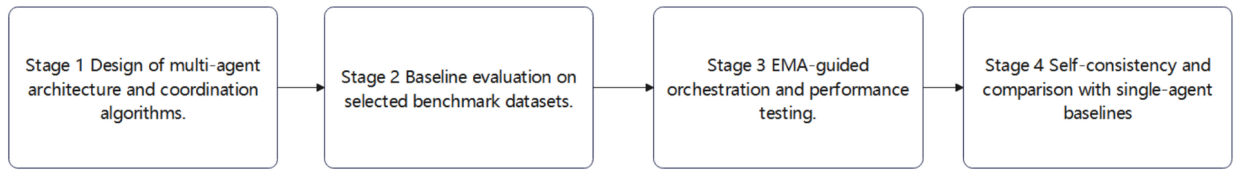

FIGURE 1
Multi-agent system architecture overview.

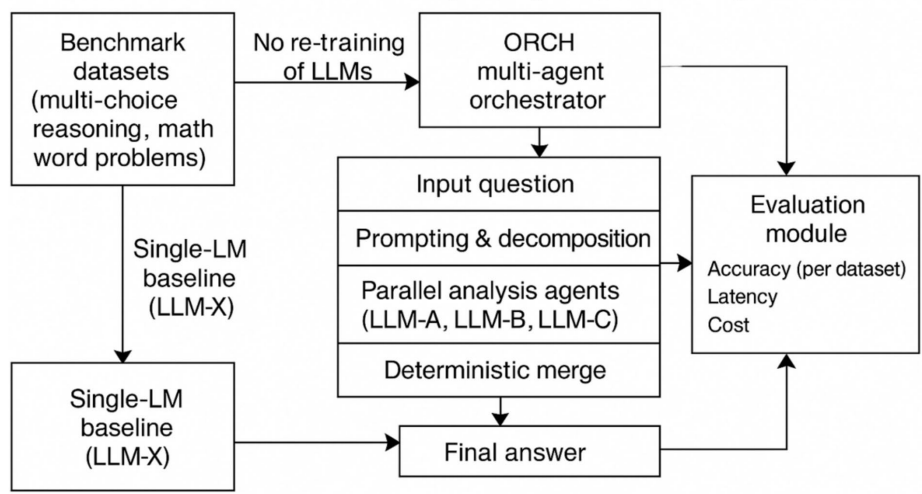

FIGURE 2
ORCH framework overall workflow.

## 3.2 Multi-agent design

Conventional LLM deployments typically follow a single-model pattern: once a request is issued, an LLM such as ChatGPT returns a direct response, either through a web interface or a client application. In non-fully-automated scenarios, many organisations still rely on human operators to intermediate such interactions. However, in domains like healthcare or financial decision-making, it is increasingly common to integrate LLMs via APIs to enable automated processing rather than manual, page-based workflows. In these settings, a key question arises: which provider and which model should be used for a given request? Relying on a single vendor LLM can be problematic, as individual models often exhibit systematic biases or insufficient accuracy on certain classes of queries. As a result, many companies opt to call multiple LLMs from different providers in parallel, which in turn creates a new challenge—how to coordinate these models so that they can collaborate effectively on the same task.

In this work, we focus on a multi-agent configuration in which three provider-specific agents are instantiated, although in principle the framework can accommodate an arbitrary number of agents. The restriction to three agents is purely for experimental convenience and clarity of analysis. All agents are orchestrated by the proposed multi-agent orchestration ORCH, which follows a "Many Analyses, One Merge" architecture: several LLM-based agents independently analyse the input, and their outputs are subsequently consolidated by a dedicated merger to produce a single, final decision as Figure 3.

The first layer is the task intake and formatting layer, which converts the original problem statement and options into a unified input template. For multiple-choice questions, the prompt explicitly requires a separate analysis for each candidate option; for open-ended problems, it instructs the model to produce a clear, step-by-step reasoning chain (Kıyak and Emekli, 2024).

The second layer is the parallel multi-agent analysis layer. Under a shared prompt template, multiple heterogeneous LLM providers (three in this study, denoted Agent-1, Agent-2, and Agent-3) are invoked in parallel. Each agent independently performs reasoning and generates its own candidate answer.

The third layer is the merger layer, which employs a dedicated merging model to compare and integrate the analysis texts and candidate answers produced by the agents, and then outputs a single final prediction.

The entire pipeline relies solely on inference-time API calls, without any parameter updates on the underlying models, and can therefore be regarded as an external orchestration and decision framework applied to black-box LLMs.

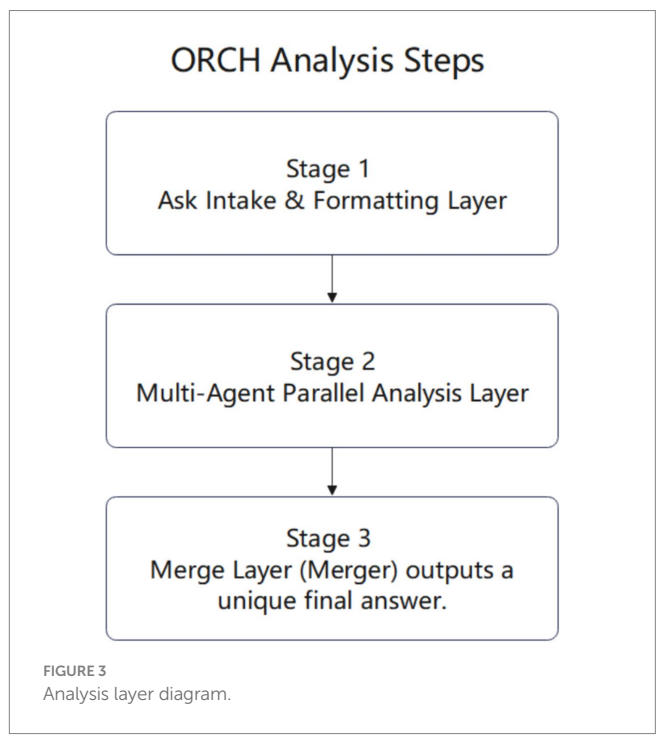

FIGURE 3
Analysis layer diagram.

TABLE 3 Benchmark dataset characteristics.

| Benchmark | Task type | Domain/content | Options/answer form | Main ability tested |
|---|---|---|---|---|
| MMLU | Multiple-choice, 4 options | 57 subjects covering humanities, social sciences, STEM, etc. | Single-answer MCQ (A–D) | Broad factual knowledge and general reasoning ability |
| MMLU-Pro | Multiple-choice, 10 options | Filtered and strengthened subset of MMLU with higher difficulty and reasoning load | Single-answer MCQ (A–J) | More challenging cross-domain reasoning and robustness to distractors |
| GSM8K | Open-ended math word problems | Grade-school/middle-school arithmetic word problems with multiple reasoning steps | Free-form numeric answer (exact match) | Multi-step mathematical reasoning and precise calculation (chain-of-thought–style reasoning) |

## 3.3 Selection of evaluation datasets

For domains such as finance and healthcare, many questions are inherently speculative and often lack a single, objectively correct answer. To avoid this ambiguity, the present study evaluates ORCH exclusively on benchmarks with well-defined gold answers for LLM-based reasoning, namely MMLU, MMLU-Pro, and GSM8K. The main characteristics of these datasets are summarised in Table 3.

These three benchmarks are chosen to examine how the "Many Analyses, One Merge" architecture of ORCH behaves across different levels and types of difficulty, considering not only accuracy but also inference time and API cost. MMLU serves as the baseline four-option multiple-choice benchmark for testing general knowledge and reasoning in LLMs. MMLU-Pro, with ten options per question and more demanding items, acts as an advanced benchmark that places greater emphasis on fine-grained discrimination and deeper reasoning. To avoid overfitting the evaluation to MMLU-style multiple-choice formats, we additionally include the open-ended mathematical benchmark GSM8K, in which answers are numerical and judged by exact match. This reduces the impact of "educated guessing" and provides a complementary perspective on ORCH ability to handle structured, multi-step reasoning.

## 3.4 ORCH with the evaluation benchmarks

Conceptually, the proposed multi-agent framework is intended as a general-purpose architecture. However, to obtain reliable and comparable empirical results, ORCH must be adapted so that its decomposition and answering procedures align with the specific formats of the chosen benchmarks.

In the present experimental setup, we take the four-option multiple-choice format of MMLU as the primary example. Before any item enters the multi-agent layer, it first passes through a standardisation step in which the original question stem and its four options are read from the dataset and reformatted into a fixed English template, such as: "Question: …/Options: A. … B. … C. … D. …." This step removes formatting differences across subjects and source files. The orchestration then invokes a designated LLM to act as a dispatcher: given the full question, this dispatcher performs a high-level analysis and automatically proposes three complementary sub-questions, typically covering perspectives such as concept verification, option elimination, and consistency checking.

If the automatic decomposition is judged unreliable, the system falls back to a predefined set of three generic sub-questions. This stage effectively inserts an explicit "subtask decomposition" layer on top of a unified template, providing structured support for downstream multi-agent collaboration.

We also looked into several different ways to break down the tasks during test runs. This included a single, large follow-up prompt and various forms of subquestions. In practice, the three-facet design used in ORCH showed more reliable sticking to the set output format and more consistent merging behavior. It also remained a practical balance concerning extra latency and cost. For this reason, we have chosen it as the default setup for the main experiments.

If an agent fails, the logs clearly show the failure. In these situations, the other two agents keep the pipeline moving. The output from the failed agent is treated as empty and left out of the merge step. As a result, the system can still provide a final answer, but the overall accuracy may drop because there are fewer valid agent votes. In our setup, we set the default fallback agent to ChatGPT (e.g., gpt-4o-mini) since it provides the best value and stable performance among the options we have.

In the parallel multi-agent analysis layer, three heterogeneous LLM providers are selected—OpenAI gpt-4o-mini, DeepSeek-chat, and XAI Grok-2-latest—denoted Model-O, Model-D, and Model-X, respectively, and each wrapped as an independent agent. For every four-option question, ORCH combines the full item with the three sub-questions to construct analysis prompts, which are then assigned to the three agents in a one-to-one manner and issued in parallel via a thread pool. Each agent is instructed, under a shared prompt template, to produce a structured analysis of its assigned sub-question in roughly 500 words, explicitly stating which options are supported or ruled out and why, and to conclude with a provisional answer letter (A/B/C/D). No multiple sampling or self-consistency procedures are used, so as to reduce stochastic noise and let performance differences primarily reflect the orchestration strategy itself. On receiving the responses, the script applies character-level truncation and regular-expression parsing, with simple fallback handling for malformed outputs, to ensure that the subsequent merging stage can consume the agent analyses robustly.

The merger layer is likewise implemented using an LLM (Tam et al., 2024), typically reusing the first model in the provider list as a dedicated merging agent. Its input consists of the normalised question together with the three agent analyses. The script concatenates these analyses into a single prompt with explicit source tags and instructs the merging agent, via a carefully designed prompt, to read all evidence, compare agreements and disagreements across agents, and then output exactly one final option letter as the decision. A parsing function extracts the first valid option letter from the returned text and records it as ORCH prediction for that item. In this way, the merger acts as a protocol-level deterministic "arbiter" that collapses three independent sub-question analyses into a single discrete choice, performing external orchestration and decision optimisation over black-box LLMs without modifying any underlying model parameters as Figure 4.

The description above uses MMLU as the running example. When the evaluation is switched to MMLU-Pro, the overall orchestration pipeline remains unchanged; the only modification is that the answer space is expanded from four options to ten, while the merger still aggregates the agents analyses into a single final choice. For GSM8K, we adopt an analogous procedure: each problem is decomposed into sub-tasks, the individual agents generate their own candidate solutions, and the merger agent performs the final synthesis to produce the definitive answer.

In our implementation, the dispatcher sends decomposed subquestions to each follow-up LLM, and each agent returns a free-form textual analysis (not a single discrete label). These three textual responses are then consumed by a dedicated merger agent, which synthesizes the final decision by extracting and reconciling evidence across agents. As a result, the merged answer is not constrained to be one of the three agents' explicit choices (and in some cases an agent may not even state a single option), which is particularly relevant for MMLU-Pro, where the merger may produce a more consistent final selection after integrating partial eliminations and supporting evidence.

This behavior differs from the VOTE baseline, where each agent returns only a discrete option and the final output is obtained by majority voting. In the rare 1–1–1 tie case (three different options), VOTE applies a protocol-level deterministic tie-break rule by selecting a predefined fallback model (e.g., gpt-4o-mini or DeepSeek-chat, chosen for cost/stability considerations); this priority can be trivially changed in the code configuration.

## 3.5 EMA-based routing mechanism

Building on the base ORCH architecture, this study adds a lightweight routing module driven by an EMA to enable dynamic selection among different LLM providers (Balla et al., 2020). For each agent, the router maintains a set of time-varying statistics: recent answer correctness (used as a quality signal), average response latency, estimated cost per request, and a stability indicator capturing timeouts, API errors, or malformed outputs. Whenever a new sample is evaluated, these statistics are not simply overwritten by the latest observation; instead, each scalar metric $s$ is updated in EMA form, so that recent performance receives higher weight while older behavior is gradually down-weighted. In other words, for the $t$-th sample we compute a new value EMA_$s(t)$ as a convex combination of the current observation and the previous EMA_$s(t-1)$, with a smoothing factor in (0, 1].

Before inference begins, the router uses the current EMA values to compute an overall score for each agent. High answer quality contributes positively to this score, while latency and monetary cost are treated as penalties; agents that have recently exhibited timeouts or abnormal outputs receive an additional negative adjustment. In the default experimental setting we still query all three agents on every question, so that results remain directly comparable to the baseline ORCH configuration. In extended experiments, we also consider variants that only dispatch the two highest-scoring agents, in order to examine the trade-off between accuracy and resource usage. In this way, the EMA router introduces a simple "memory" of historical behavior and allows the call frequency of each agent to be adjusted adaptively, without changing the internal reasoning pipeline of ORCH, with the aim of achieving a more favourable balance between performance and cost. The formula is shown in Equation 1.

$$\text{EMA}_s(t) = \alpha \cdot s(t) + (1-\alpha) \cdot \text{EMA}_s(t-1) \quad (1)$$

Concretely, the EMA module tracks four indicators per agent: (i) ema_quality, where recent answers are encoded as 1 for correct and 0 for incorrect; (ii) ema_latency, reflecting typical API response time; (iii) ema_cost, representing the estimated cost of a single request; and (iv) ema_stability, summarising the incidence of timeouts, errors, or format violations. These metrics help the router decide which agent is better suited to act as the problem-decomposition dispatcher and which is more reliable as the merging agent. However, this mechanism is inherently feedback-dependent: EMA updates require knowledge of whether the last prediction was correct. As a result, EMA-based routing is naturally applicable to benchmarks with ground-truth labels, such as the MMLU family and GSM8K, but is less straightforward to use in real-world financial or medical scenarios where explicit correctness feedback is not consistently available.

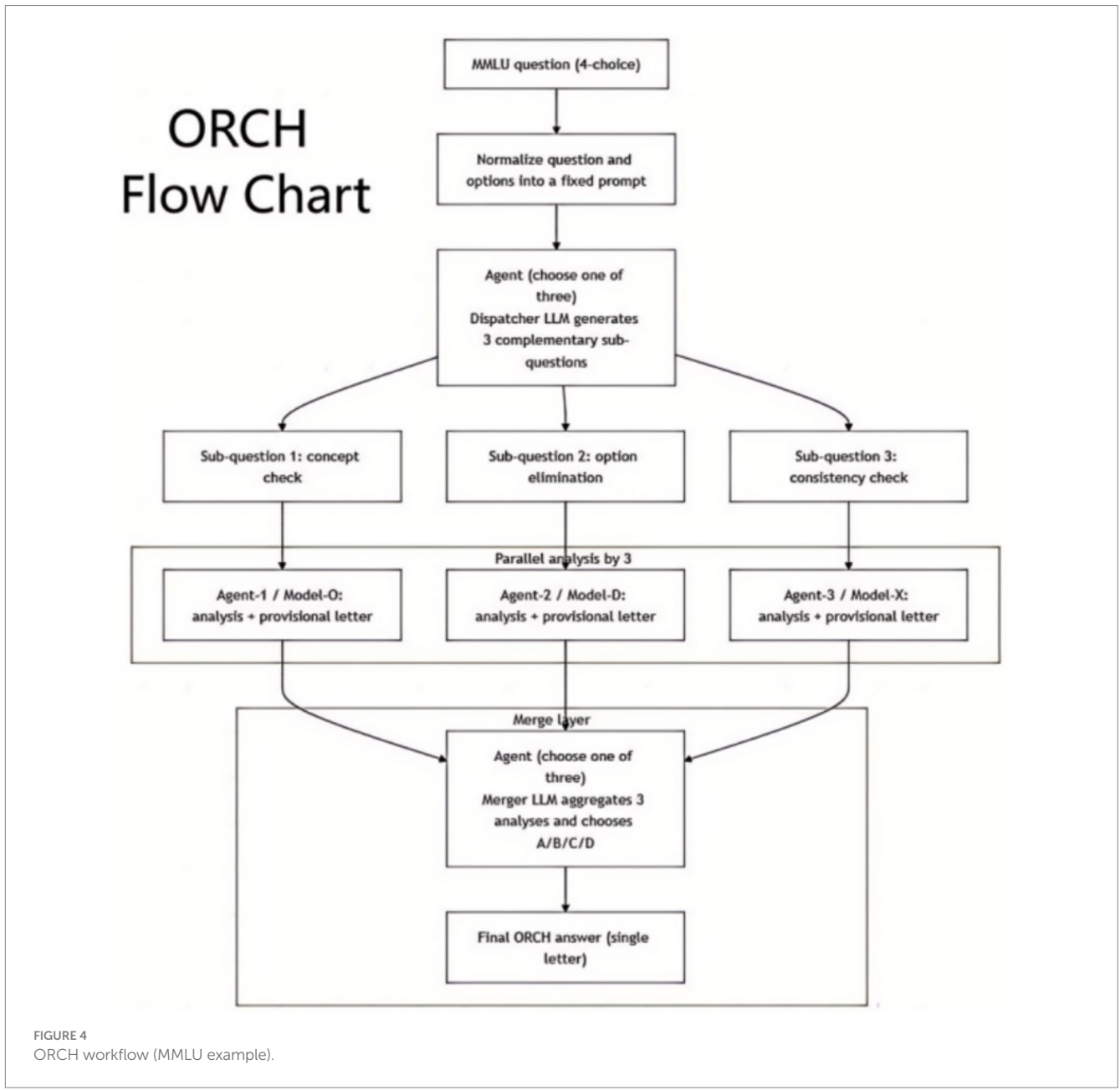

FIGURE 4
ORCH workflow (MMLU example).

EMA routing is not part of the ORCH core; it is an optional, evaluation-oriented module. In practical deployment, where gold-label feedback is unavailable, ORCH would need alternative proxies (e.g., verifier/critic scores or delayed user/human feedback) to update routing decisions.

## 3.6 Self-consistency and multi-shuffle modules

In many recent studies on multi-agent and chain-of-thought reasoning (Chen X. et al., 2023), self-consistency has become a commonly used technique. In essence, self-consistency answers the same question multiple times with a stochastic decoding strategy and then aggregates these responses—typically via majority voting or by fusing several reasoning traces—so as to obtain a more stable final conclusion. Make explicit that it intentionally uses stochastic decoding and relaxes strict determinism when enabled.

Multi-shuffle addresses a different source of instability: sensitivity to option order. For a given multiple-choice question, the options are repeatedly permuted and the model is queried on each permuted version; the resulting predictions are then combined to mitigate position bias and make the model's choice more robust to the presentation order.

In the comparative experiments of this study, we incorporate both self-consistency and multi-shuffle as optional modules. Specifically, we set the self-consistency parameter to $K = 2$, meaning that two stochastic samples are generated per query. To control computational cost, self-consistency is applied only at the merger layer, rather than at every agent, and we enable a light-weight multi-shuffle configuration with $m = 1$ is evaluated under a single shuffled option order.

## 3.7 Experimental conditions and parameter settings

The experimental environment here uses Python scripts, and all agents use API interfaces. All test parameters and conditions and model at Tables 4, 5.

### 3.7.1 MMLU

We used a fixed 10-subject subset—Abstract Algebra, Anatomy, Business Ethics, Clinical Knowledge, College Mathematics, Computer Security, Econometrics, Jurisprudence, Machine Learning, and Moral Scenarios. With seed 42, we performed uniform sampling without replacement within each subject, selecting $n = 30$ questions per subject (total $N = 300$). The sampled questions were then pooled and globally shuffled under the same seed to mitigate order effects, and the resulting indices were frozen to ensure exact reproducibility. Accordingly, all MMLU-based analyses reported in the manuscript, including EMA-style routing and self-consistency, are conducted on this identical MMLU instance set, supporting both cost-controlled experimentation and reproducible comparisons.

### 3.7.2 MMLU-Pro

Because the 10-option format (A–J) is more challenging and closer to practical evaluations, and to avoid item cherry-picking, we did not pre-specify subject subsets. Instead, we adopted a random-yet-reproducible protocol: we first restricted the test split to items with exactly 10 options, then deterministically selected the three categories with the largest eligible pools, and finally sampled 20 questions per category without replacement after shuffling each category pool under seed 42. Evaluation proceeded in category order from the first to the third selected category, with randomized ordering within each category, and all sampled indices were stored to enable replication.

### 3.7.3 GSM8K

With seed 42, we sampled $10 \times B$ test problems without replacement (default $B = 30$), sorted the sampled instances by question token length, and partitioned them into 10 equal-sized length buckets from shortest to longest. Bucket membership and within-bucket ordering were fixed to support fair length-stratified comparisons and reproducible analysis.

To reduce sampling variability and enable paired significance testing, we fix the evaluation random seed so that the sampled instance set (and its evaluation order) remains constant. This ensures that observed differences primarily reflect method or configuration changes rather than re-sampling noise, allowing paired comparisons on identical instances and the use of paired tests such as McNemar's test.

At inference time, we set temperature = 0 to minimize decoding stochasticity. However, a fixed seed and temperature = 0 mainly stabilize the client-side evaluation protocol and decoding configuration; outputs from provider-hosted services may still drift due to model-version changes or backend updates.

The experimental design is structured as follows. Starting from the choice of benchmarks, MMLU is used as one of the primary testbeds because it is relatively large while its individual items are comparatively straightforward. Considering time and computational cost, ten subjects are selected, with 30 questions per subject, yielding a total of 300 items. Preliminary runs indicate that using more than 300 questions does not lead to materially different aggregate results, so this size is sufficient for our purposes. MMLU-Pro, as a strengthened variant of MMLU, serves to more clearly reveal the effect of ORCH under higher difficulty. To avoid the risk that the proposed framework only performs well on the MMLU family, we additionally include GSM8K, which allows us to demonstrate the broader applicability of ORCH beyond standard multiple-choice settings.

With respect to comparison conditions, the first and most direct baseline is a single-agent setup, against which we compare the accuracy of ORCH. We also report a vote baseline, where the final answer is determined by majority voting over the three AI agents; If they are still the same, break the tie using a fixed priority order (e.g., O > D > X) or alphabetical order.

In the ablation study, we remove one agent and examine the two-agent variant (denoted ORCH-1) to assess how reducing the number of agents affects test accuracy. A further comparative setting introduces the EMA module to see whether EMA-guided routing yields improvements over the vanilla ORCH. The final comparison adds the self-consistency module to evaluate its impact on performance.

Regarding evaluation metrics, we report not only the accuracy of ORCH itself, but also the standalone accuracies of the three individual agents and the vote baseline. In addition, we analyse latency and API cost, and we apply the McNemar test to assess whether observed accuracy differences between systems are statistically significant.

Code availability: The code for reproducing the experiments and running the pipeline is publicly available at: https://github.com/mmzr123/ORCH-zhouhanlin.

TABLE 4 Test model and dataset.

| Test set model | MMLU | MMLU-PRO | GSM8K |
|---|---|---|---|
| Number of problems | There are 30 questions in each of the 10 subjects, and 300 questions in each dataset. | | There are 30 questions each for B1–B10, for a total of 300 questions. |
| Agent 1 | OpenAI gpt-4o-mini | | |
| Agent 2 | DeepSeek-chat | | |
| Agent 3 | XAI Grok-2-latest | | |

TABLE 5 Test comparison conditions and indicators.

| Comparison condition | | |
|---|---|---|
| Comparison condition 1: ORCH (3 test sets) | ORCH compared with three single-agent baselines | Test metrics: 1. ORCH accuracy 2. Baseline accuracy of 3 single agents 3. VOTE accuracy 4. Single subject accuracy 5. Latency, estimated cost 6. MCNEMAR test |
| Comparison condition 2 ORCH-1 | Ablation experiment, removing one agent (XAI) | |
| Comparison condition 3 | Add EMA module | |
| Comparison condition 4 | Add a self-consistent module $K = 2$ | |

# 4 Experiment result and analysis

## 4.1 Comparative experiment 1–1 ORCH vs. three baselines (MMLU)

In this setting, the three baselines correspond to using each model independently: OpenAI gpt-4o-mini, DeepSeek-chat, and XAI Grok-2-latest. The vote baseline is defined as follows: if at least two of the three baselines produce the same answer, that answer is selected; if all three disagree, one of the three answers is chosen at random. The experimental results are summarised as follows: Figure 5 reports the global accuracy, while Figure 6 presents the per-subject accuracy, and Figure 7 shows the latency comparison across all methods. Subsequent significance testing will follow the same procedure as in Figure 8. In the following sections, we will not report all confusion matrices.

Because the token prices of the three provider APIs differ, we approximate the relative cost as follows: if O, D, and X denote the per-call token cost of OpenAI gpt-4o-mini, DeepSeek-chat, and XAI Grok-2-latest, respectively, then the cost of VOTE is roughly O + D + X, whereas ORCH, which queries three agents plus a merger, is approximately 3 × (O + D + X).

On the MMLU four-option benchmark (10 subject subsets, 300 questions), ORCH is compared against three single-model baselines (OpenAI gpt-4o-mini, DeepSeek-chat, XAI Grok-2-latest) and a simple majority-vote strategy (VOTE). As shown in Figure 5 (Global Accuracy), ORCH achieves the highest overall accuracy (0.813), outperforming OpenAI (0.683), DeepSeek (0.730), XAI (0.767), and VOTE (0.740). This corresponds to an absolute gain of 4.6 percentage points over the strongest single model (XAI), 7.3 points over VOTE, and 8.3–13.0 points over OpenAI and DeepSeek. Figure 6 (Per-subject Accuracy) further shows that ORCH attains the best performance on most subjects—including abstract_algebra, anatomy, clinical_ knowledge, college_mathematics, jurisprudence, and machine_ learning—and remains close to the top method on the remaining subjects, indicating good robustness and cross-domain generalisation.

In terms of efficiency, Figure 7 (Latency) shows that ORCH (three agents plus merger) has an average latency of 11,775 ms per question, compared with 853, 1,558, and 622 ms for OpenAI, DeepSeek, and XAI, and 3,033 ms for VOTE. Thus ORCH trades roughly three times the cost and about four times the latency of a simple three-model vote for a 4–13 point accuracy improvement. This accuracy–cost trade-off is attractive for high-value decision or expert QA scenarios where accuracy is paramount, whereas XAI alone or VOTE may be preferable in real-time or cost-sensitive applications compare data at Table 6.

It should be noted that, in terms of approximate per-token API cost, XAI is the most expensive, followed by DeepSeek, with OpenAI gpt-4o-mini being the cheapest. Compared with OpenAI and DeepSeek, the accuracy gains achieved by ORCH are statistically significant at the 5% level. When contrasted with the strongest single-model baseline, XAI, ORCH answers 14 more questions correctly out of 300; the corresponding McNemar test yields $\chi^2 \approx 2.91$ and $p \approx 0.088$, while not reaching the traditional 95% significance level, thus indicating that ORCH has only a slight numerical advantage in this dataset and scenario.

## 4.2 Comparative experiment 1–2: ORCH vs. three baselines (MMLU-Pro)

In this experiment, we evaluate all methods on the MMLU-Pro dataset (Xuan et al., 2025). The three single-model baselines and the VOTE strategy are defined in the same way as in Section 4.1.1. The results are summarised as follows: Figure 9 reports the global accuracy, Figure 10 presents the per-subject accuracy, and Figure 11 shows the latency comparison across all methods.

On the MMLU-Pro benchmark (10-choice, 10 subject areas), ORCH clearly outperforms all three single-model baselines and the simple majority-vote strategy. Its overall accuracy reaches 0.733, compared with 0.557 for DeepSeek-chat, 0.450 for OpenAI gpt-4o-mini, 0.510 for XAI Grok-2-latest, and 0.507 for VOTE, corresponding to absolute gains of 17.6, 28.3, 22.3, and 22.6 percentage points, respectively. At the subject level, ORCH attains the best or tied-best accuracy in most domains, including physics, math, chemistry, engineering, business, law, and economics; it is only slightly behind DeepSeek or VOTE on psychology and biology (Bach, 2025), and the gaps there remain small.

This performance improvement comes at a higher computational cost. The average per-question latency of ORCH is about 18,263 ms, substantially larger than that of the single models (≈617 ms for XAI, ≈1,010 ms for OpenAI, ≈1,387 ms for DeepSeek) and VOTE (≈3,013 ms). Assuming per-call token costs D, and X for OpenAI gpt-4o-mini, DeepSeek-chat, and XAI Grok-2-latest, the cost of VOTE is roughly O + D + X, whereas ORCH is approximately 3 × (O + D + X), as in the MMLU setting.

For significance analysis, we apply the McNemar test comparing ORCH with the strongest single-model baseline, DeepSeek-chat. Among 300 questions, we obtain $b = 69$ cases where ORCH is correct and DeepSeek is wrong, and $c = 16$ cases with the opposite pattern, so $b + c = 85$ items differ in correctness. The continuity-corrected McNemar statistic is $\chi^2 = 33.0471$ with an associated $p \approx 8.99 \times 10^{-9}$, far below 0.05, indicating that the performance gain of ORCH over DeepSeek-chat is statistically significant at the 95% confidence level.

## 4.3 Comparative experiment 1–3: ORCH vs. three baselines (GSM8K)

In this experiment, all methods are evaluated on the GSM8K dataset. The three single-model baselines and the VOTE strategy are defined in the same way as in the previous subsections. The results are summarised as follows: Figure 12 reports the global accuracy, Figure 13 presents the per-category accuracy, and Figure 14 shows the latency comparison across all methods.

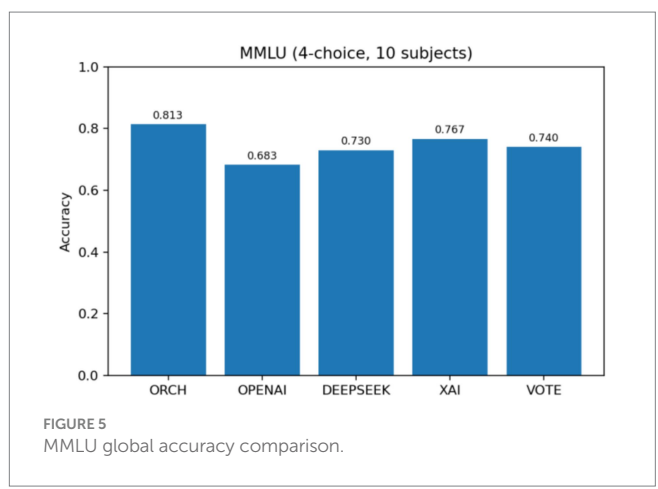

FIGURE 5
MMLU global accuracy comparison.

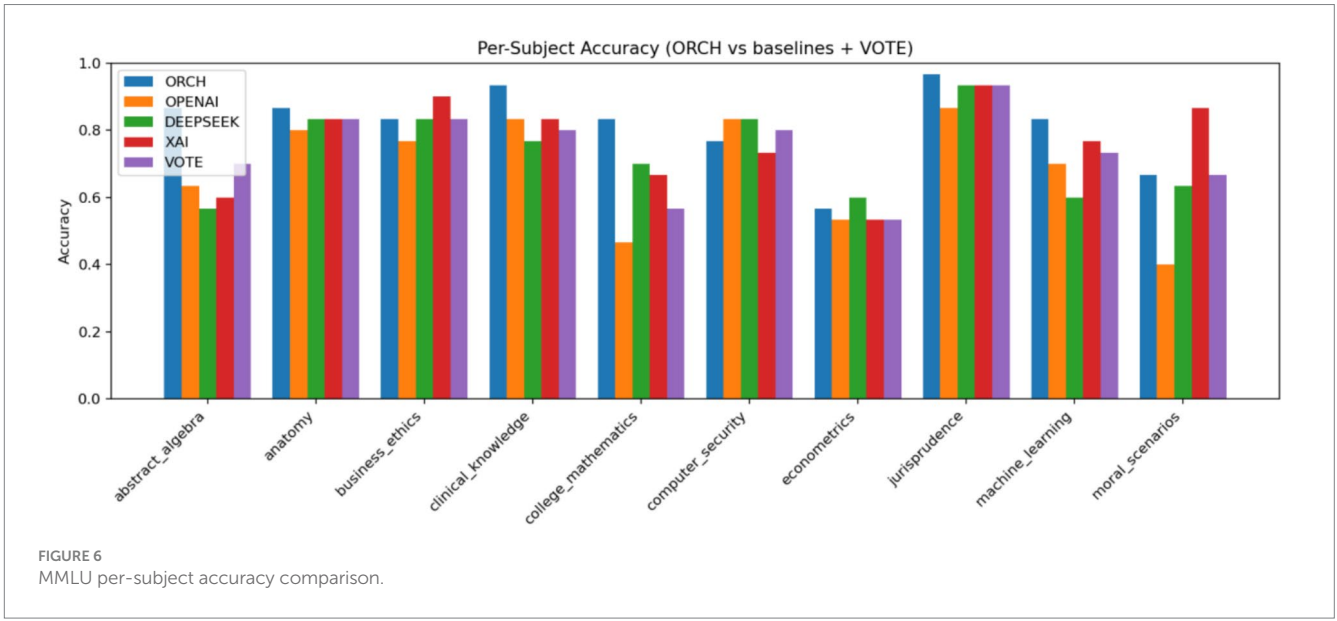

FIGURE 6
MMLU per-subject accuracy comparison.

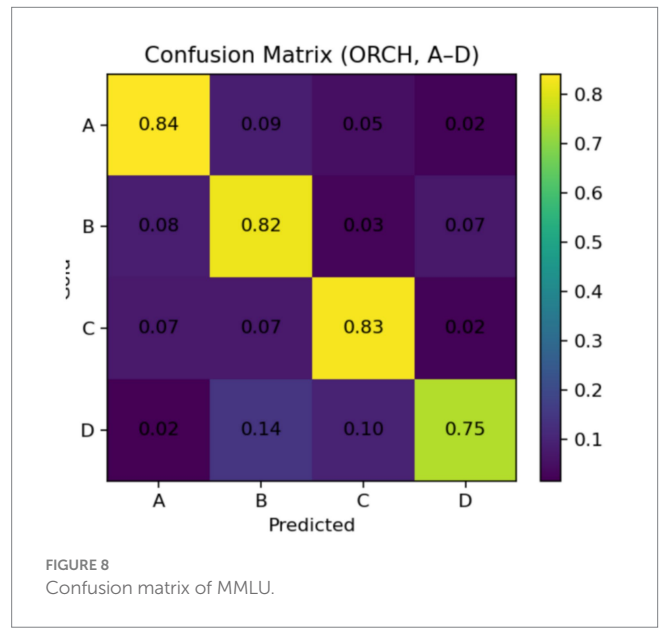

FIGURE 7
MMLU latency comparison.

FIGURE 8
Confusion matrix of MMLU.

On the GSM8K mathematical reasoning benchmark (10 length buckets × 30 questions), ORCH markedly outperforms all three single-model baselines and the VOTE strategy. Its global accuracy reaches 0.940, whereas OpenAI gpt-4o-mini, DeepSeek-chat, XAI Grok-2-latest, and VOTE achieve only 0.277, 0.440, 0.413, and 0.397, respectively. Across all 10 length buckets, ORCH maintains accuracy at or above 0.9—even for longer, multi-step problems—while the baselines and VOTE typically remain in the 0.2–0.6 range, indicating that the multi-agent decomposition-and-merge scheme is particularly effective for multi-step numerical reasoning.

This gain comes with higher computational cost: the average latency per question is about 11,453.5 ms for ORCH, compared with roughly 618.5 ms (XAI), 989.1 ms (GPT-4o-mini), 1,355.6 ms (DeepSeek), and 2,963.3 ms (VOTE). If the per-call token costs of OpenAI, DeepSeek, and XAI are denoted by O, D, and X, then VOTE costs approximately O + D + X, whereas ORCH, due to multiple calls in both analysis and merging stages, is about 3 × (O + D + X). In other words, ORCH trades roughly three times the token cost of VOTE and substantially higher latency for a much larger accuracy improvement.

McNemar's test confirms that this improvement is not due to random fluctuation: when comparing ORCH with the strongest single-model $c = 1$ (i.e., 151 questions answered correctly only by ORCH and 1 only by DeepSeek), yielding $\chi^2 \approx 148.03$ and $p \approx 8.99 \times 10^{-34}$ <0.05, indicating a highly significant difference at the 95% confidence level.

## 4.4 Comparative experiment 2—ORCH-1 ablation: removing one agent (XAI)

To verify that the performance gains of ORCH indeed stem from the cooperation of multiple agents, we conduct an ablation study in which one agent (XAI) is removed, yielding a two-agent variant denoted ORCH-1 (Levy et al., 2024). We then compare ORCH-1 against the full ORCH to assess how much performance changes when one agent is dropped. In this experiment, we use the MMLU-Pro dataset: XAI is disabled only in ORCH-1, while the baseline runs and the VOTE strategy still include XAI as in previous sections. The results are summarised as follows: Figure 15 reports the global accuracy, Figure 16 shows the per-subject accuracy, and Figure 17 presents the latency comparison.

TABLE 6 ORCH vs. three baselines (MMLU).

| Method | Type | Global accuracy | Average latency (ms/question) | Description |
|---|---|---|---|---|
| ORCH | Multi-agent orchestration (3-agent + merge) | 0.813 | 11,775 | Highest accuracy, highest cost and latency |
| XAI | Single model baseline | 0.767 | 622 | Strongest single-model baseline |
| DeepSeek | Single model baseline | 0.730 | 1,558 | Accuracy second to XAI, medium latency |
| OpenAI | Single model baseline | 0.683 | 853 | Lowest accuracy, relatively low latency |
| VOTE | Multi-model majority voting (3-model) | 0.740 | 3,033 | Mid-level accuracy, moderate cost and latency |

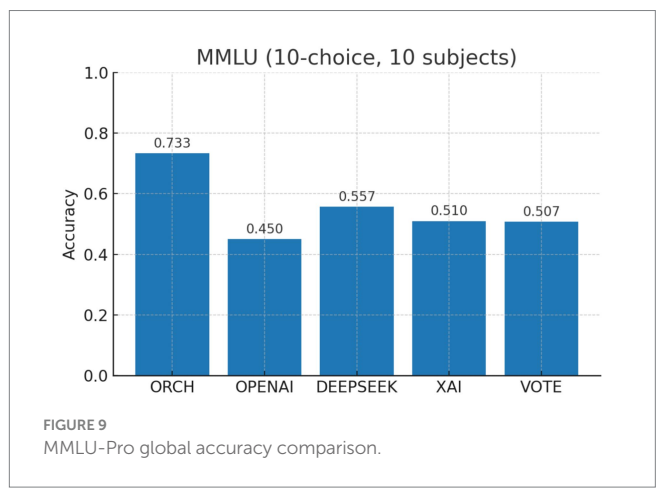

FIGURE 9
MMLU-Pro global accuracy comparison.

On MMLU-Pro, the ablation where one agent (XAI) is removed shows that ORCH-1 attains a global accuracy of 0.700, lower than the 0.733 of the full three-agent version in Section 4.1.2, but still clearly above all single-model baselines (OpenAI 0.460, DeepSeek 0.560, XAI 0.507) and the VOTE scheme (0.513). At the subject level, ORCH-1 continues to achieve the best or tied-best accuracy in most areas, including physics, math, chemistry, business, and law, although it lags the three-agent ORCH by about 2–5 percentage points in several subjects (e.g., math, chemistry, other, psychology, biology), indicating that XAI as a third agent contributes to cross-domain robustness. In terms of efficiency, the average latency of ORCH-1 is about 16873.8 ms per question, slightly below the roughly 18,000 ms of the three-agent setting but still far higher than the single models (XAI ≈ 626 ms, GPT-4o-mini ≈ 1,055 ms, DeepSeek ≈ 1,395 ms) and VOTE (≈ 3,076.6 ms). With O and D denoting the per-call token costs of OpenAI and DeepSeek, the cost of ORCH-1 can be approximated as ≈2.5 × (O + D), while VOTE remains O + D + X. A McNemar test comparing ORCH-1 with the strongest single model, DeepSeek-chat, yields $b = 66$, $c = 24$, $\chi^2 \approx 19.60$, and $p \approx 9.55 \times 10^{-6} < 0.05$, confirming that the improvement of ORCH-1 over DeepSeek-chat is still statistically significant. Overall, removing one agent leads to a noticeable but moderate degradation, supporting that the performance of the full three-agent ORCH stems from genuine cooperative effects rather than from any single strongest baseline.

Below is a comparison between ORCH and ORCH-1, as shown in Figure 18.

## 4.5 Comparative experiment 3—ORCH with EMA routing

In this setting, the EMA-based routing module is enabled. For each new question, the router updates agent scores according to the previous item's correctness and jointly considers latency, estimated cost, and recent stability to decide which agent serves as the decomposition (dispatcher) layer and which acts as the merger. The evaluation results are summarised as follows: Figure 19 reports the global accuracy on MMLU-Pro, Figure 20 shows the corresponding MMLU-Pro latency, Figure 21 presents the global accuracy on MMLU-Pro, and Figure 22 displays the latency on MMLU-Pro.

On the MMLU four-option, multi-subject benchmark, enabling EMA-based routing raises the global accuracy of ORCH to 0.820, clearly above the three single-model baselines (0.697 for OpenAI gpt-4o-mini, 0.723 for DeepSeek-chat, 0.750 for XAI Grok-2-latest) and the VOTE strategy (0.737). The gain over the strongest single model, XAI, is roughly 7 percentage points, indicating that dynamically selecting which agent serves as the decomposition agent and which as the merger can better exploit complementarities among models. This improvement, however, comes with higher computational cost: the average latency per question for ORCH is about 12,315 ms, substantially larger than that of the individual models (XAI ≈ 659 ms, OpenAI ≈ 954 ms, DeepSeek ≈ 1,425 ms) and VOTE (≈3,037 ms), reflecting a typical trade-off of higher accuracy for increased delay. McNemar's test ($b = 44$, $c = 18$, $\chi^2 \approx 10.08$, $p \approx 0.0015$) confirms that the advantage of ORCH over XAI is statistically significant at the 95% confidence level, rather than a result of random fluctuation.

On the more challenging MMLU-Pro benchmark (10 options, 10 subject areas), ORCH with EMA routing attains a global accuracy of 0.753, clearly surpassing OpenAI gpt-4o-mini (0.453), DeepSeek-chat (0.563), XAI Grok-2-latest (0.510), and the VOTE baseline (0.510). Relative to the strongest single model, DeepSeek, this corresponds to an improvement of about 19 percentage points, and ORCH still holds an advantage of roughly 24 points over VOTE, indicating that dynamic routing yields particularly strong multi-agent gains when options are numerous and reasoning chains are longer. This comes at a substantial computational cost: the average latency of ORCH rises to 19,645.7 ms per question, more than six times that of VOTE (≈3,203 ms) and far above the single models (XAI ≈ 646.5 ms, GPT-4o-mini ≈ 1,181.6 ms, DeepSeek ≈ 1,375.3 ms), confirming that ORCH remains a high-cost configuration on MMLU-Pro. McNemar's test ($b = 75$, $c = 18$, $\chi^2 \approx 34.94$, $p \approx 3.41 \times 10^{-9}$) further shows that the improvement of ORCH over DeepSeek is highly significant statistically, providing strong evidence for the effectiveness of EMA routing on this benchmark.

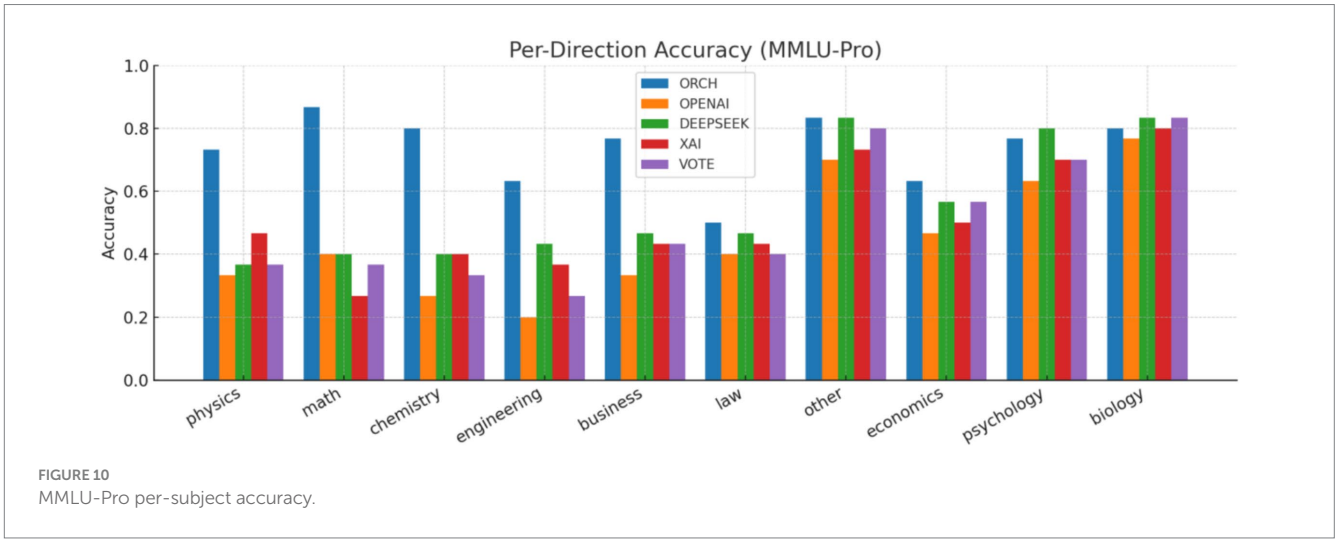

FIGURE 10
MMLU-Pro per-subject accuracy.

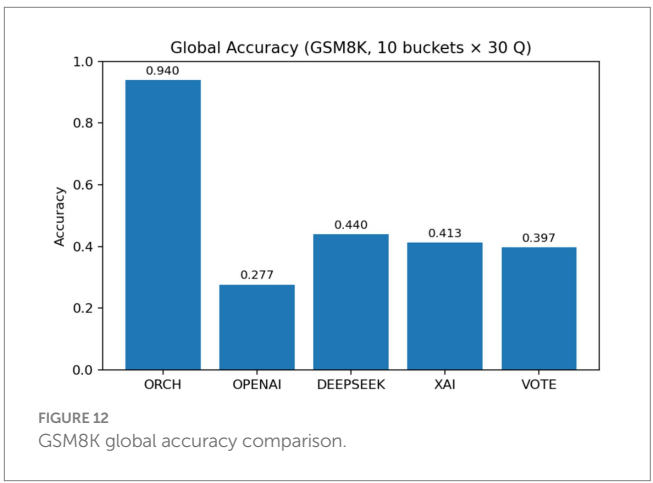

FIGURE 11
MMLU-Pro latency analysis.

FIGURE 12
GSM8K global accuracy comparison.

### 4.5.1 Overall trends and cost trade-offs on the two benchmarks

Taken together, the MMLU and MMLU-Pro results reveal a consistent pattern. On both benchmarks, ORCH with EMA routing clearly outperforms the three single-model baselines and the simple majority-vote scheme. On MMLU, the gain is relatively moderate (about 7 percentage points over the strongest baseline) and is reflected mainly in more stable performance across subjects and fewer catastrophic errors. On MMLU-Pro, the improvement is much larger (around 19 percentage points over the strongest baseline), suggesting that in "many options, high complexity, long-chain reasoning" settings, dynamically deciding which agent handles decomposition and which performs merging is particularly important. At the same time, ORCH incurs substantially higher latency than both single models and VOTE on both tasks; the EMA-enhanced variant can thus be viewed as the "high-accuracy, high-cost" corner of the design space on MMLU/MMLU-Pro, suitable for applications where answer quality is critical and real-time or cost constraints are relatively relaxed.

Compared with the fixed-routing version of ORCH (without EMA), enabling EMA routing increases accuracy from 0.813 to 0.820 on MMLU and from 0.733 to 0.753 on MMLU-Pro—a modest gain of about 0.7–2 percentage points, but consistent and reproducible across both benchmarks. The corresponding average latency rises from 11,775 ms to 12,315 ms on MMLU and from 18,263.4 ms to 19,645.7 ms on MMLU-Pro, roughly 5–10% additional inference time and token cost. This configuration is therefore most appropriate in scenarios that prioritise accuracy and robustness over strict real-time requirements as Table 7.

While ORCH attains statistically significant gains over the best single-model baseline under the chosen evaluation setting, the incremental effect of adding EMA routing within ORCH is not statistically significant on either MMLU or MMLU-Pro. On the MMLU four-option benchmark ($N = 300$), we used McNemar's paired test to compare ORCH with EMA routing against the fixed-routing version. The contingency counts are $b = 15$ (items answered correctly only by EMA–ORCH) and $c = 13$ (items answered correctly only by fixed ORCH), so $b + c = 28$, yielding $\chi^2 \approx 0.14$ and $p \approx 0.71$. With a significance level of $\alpha = 0.05$, this result does not allow us to statistically distinguish the two systems, even though the overall accuracy increases slightly from 0.813 to 0.820. The analysis on MMLU-Pro is analogous: $p$-values remain above 0.05, while the accuracy point estimates consistently favor the EMA-based variant, suggesting a small but directionally consistent improvement under the selected models, prompts, and fixed sampled question sets, rather than a strong or general advantage.

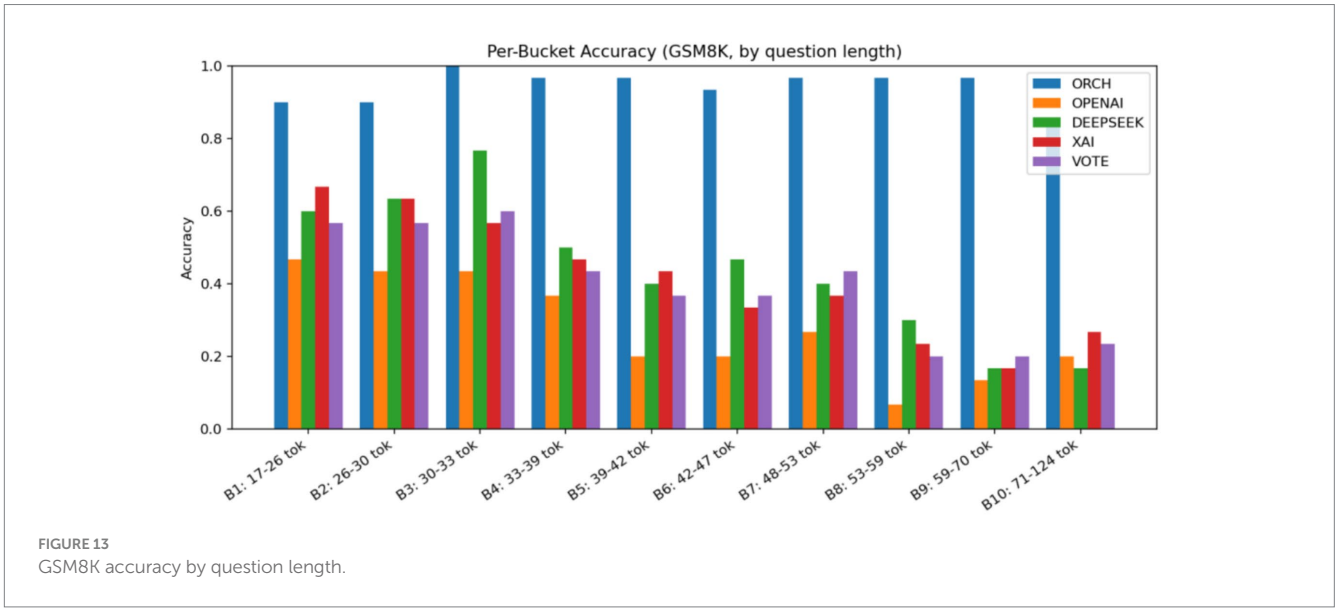

FIGURE 13
GSM8K accuracy by question length.

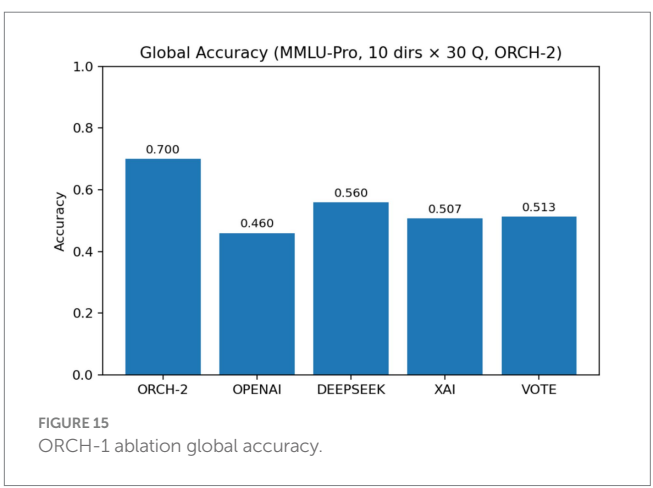

FIGURE 14
GSM8K latency analysis.

FIGURE 15
ORCH-1 ablation global accuracy.

## 4.6 Comparative experiment 4—adding the self-consistency module

In this comparative setting, we additionally enable the self-consistency and multi-shuffle modules (Wang et al., 2025). Specifically, the self-consistency parameter is set to $K = 2$, i.e., each item is answered two times under stochastic sampling. To contain the computational cost, self-consistency is applied only at the merger layer. In the same layer, we also use a light-weight multi-shuffle configuration with $m = 1$, meaning that the options are permuted once per question before aggregation. The resulting global accuracies are reported. Figure 23 presents the global accuracy on MMLU-Pro, Figure 24 presents the per-category accuracy, and Figure 25 shows the latency comparison.

In comparative setting 3, we augment the ORCH framework with a self-consistency module ($K = 2$, i.e., two stochastic samples) and a multi-shuffle mechanism ($m = 1$), and apply self-consistent reasoning plus a single option shuffling step only at the merge layer. On MMLU-Pro (10 subject areas), the resulting ORCH variant attains a global accuracy of 0.740, still clearly above the three single-model baselines (OpenAI 0.443, DeepSeek 0.563, XAI 0.507) and the VOTE scheme (0.497). At the subject level, ORCH remains markedly ahead in math, chemistry, other, psychology, and biology, and is generally comparable to or better than the strongest single model in physics, business, law, and economics, suggesting that self-consistency plus multi-shuffle provides some additional robustness for the multi-agent merger on complex, many-option reasoning tasks.

The price, however, is an extremely large increase in computational cost: the average per-question latency of ORCH rises to 89,462 ms, far exceeding VOTE (≈3,051 ms) and all single models (DeepSeek ≈1,366 ms, GPT-4o-mini ≈1,039 ms, XAI ≈ 646 ms), placing this configuration at the "ultra-high-cost" end of the spectrum. Compared

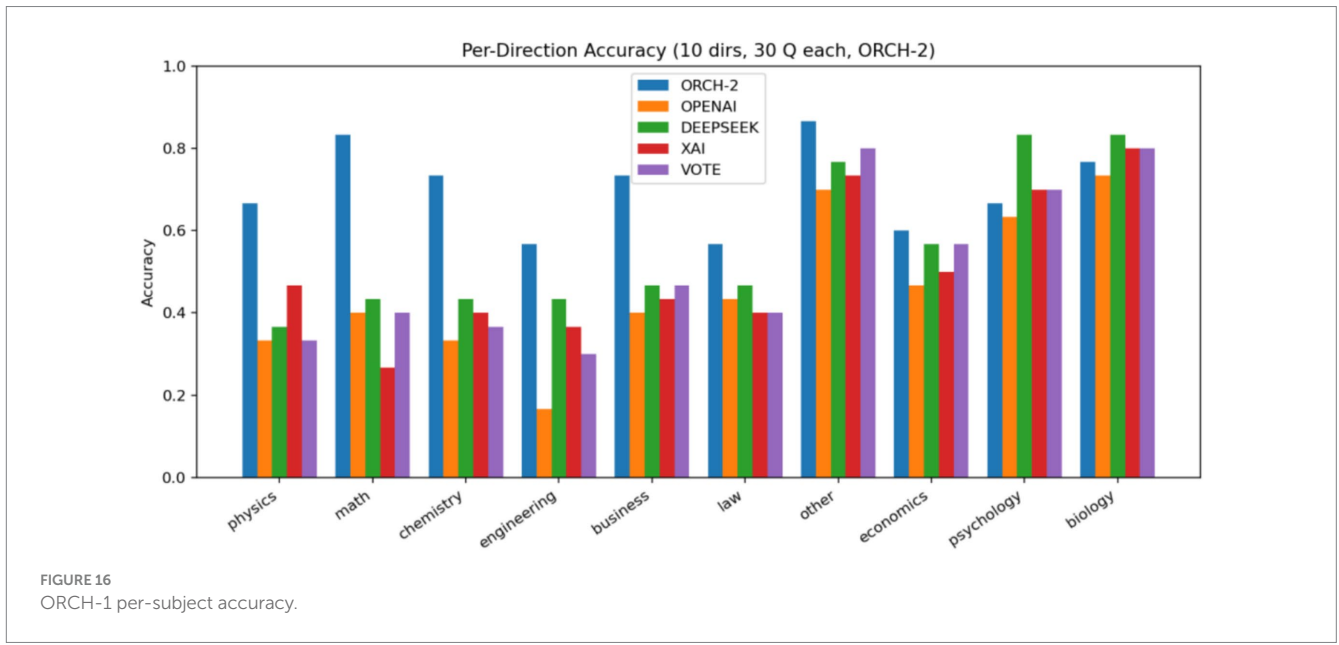

FIGURE 16
ORCH-1 per-subject accuracy.

| ORCH-1 Latency | |
|---|---|
| Model | Avg latency (ms) |
| ORCH-1 | 16873.8 |
| VOTE (3-model) | 3076.6 |
| DEEPSEEK | 1395.2 |
| GPT-4o-mini | 1055.1 |
| XAI (grok) | 626.3 |

FIGURE 17
ORCH-1 latency comparison.

## 4.7 Chapter summary

Across the three benchmarks MMLU, MMLU-Pro, and GSM8K, the multi-agent orchestration model ORCH consistently outperforms all three single-model baselines (GPT-4o-mini, DeepSeek-chat, XAI Grok-2) as well as the simple majority-vote scheme VOTE, despite the differences in task format, difficulty, and dataset. On the relatively easier four-option MMLU setting, ORCH achieves roughly a 4–7 percentage point gain over the strongest single model XAI. On the more demanding ten-option, multi-domain MMLU-Pro benchmark, the improvement increases to about 17–19 percentage points. For GSM8K, which focuses on numerical reasoning, ORCH attains an accuracy of 0.94, compared with approximately 0.44 for the best single model, highlighting the clear advantage of the multi-agent decomposition-and-merge mechanism on complex, multi-step reasoning tasks. McNemar tests further show that, on MMLU-Pro and GSM8K, the gains of ORCH over the strongest single baseline (DeepSeek) reach or substantially exceed the 95% confidence level under multiple experimental configurations, indicating that these improvements are statistically robust rather than random fluctuations.

The ablation and variant studies provide additional insight into the contribution of individual components within the ORCH framework. Removing one agent (ORCH-1, retaining only OpenAI and DeepSeek) leads to an accuracy drop of around 3 percentage points, while latency and cost decrease only modestly, suggesting that the three-agent configuration offers meaningful diversity and complementarity beyond any single strong baseline. The EMA routing module yields a small but consistent accuracy gain of about 0.7–2.0 percentage points on MMLU and MMLU-Pro, at the expense of roughly 5–10% extra latency and token cost: the improvement over single models is statistically significant, whereas the direct paired comparison with fixed-routing ORCH on MMLU does not yet cross the usual significance threshold, manifesting more as a stable "better on average" trend. By contrast, adding self-consistency ($K = 2$) and multi-shuffle ($m = 1$) on top of ORCH+EMA still keeps performance clearly above all single models and VOTE, but slightly reduces with the EMA-only ORCH on MMLU-Pro (global accuracy 0.753, average latency ≈19,645.7 ms), adding self-consistency and multi-shuffle actually reduces accuracy slightly (0.753 → 0.740, about −1.3 percentage points) while increasing latency by more than a factor of four (19.6 s → 89.5 s per question). Under the current settings ($K = 2$, $m = 1$, self-consistency only at the merge layer), the self-consistency module therefore does not yield additional accuracy gains and instead dramatically inflates inference cost. From a cost–effectiveness perspective, this variant is better viewed as an upper-bound exploration demonstrating that ORCH continues to dominate the baselines under very high compute budgets, rather than as a practical default configuration for deployment.

Table 8 indicates that the consistency module is not strictly necessary in the ORCH framework. On the one hand, enabling this module raises the token cost of ORCH to approximately $6 \times (O + D + X)$ and increases latency to more than four times that of the version without it; on the other hand, accuracy not only fails to improve but actually declines.

**FIGURE 18**
ORCH-1 vs. ORCH.

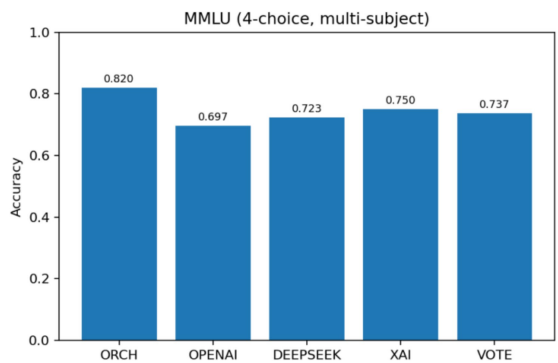

**FIGURE 19**
EMA routing MMLU-Pro global accuracy.

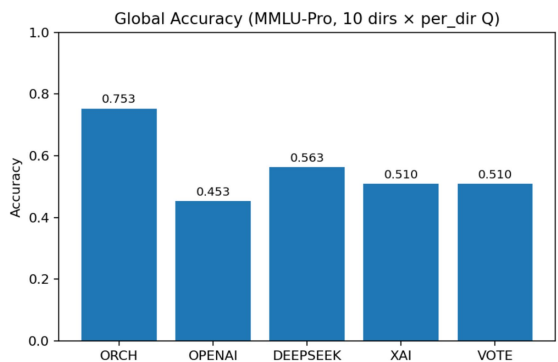

**FIGURE 20**
EMA routing MMLU-Pro latency.

**FIGURE 21**
EMA routing MMLU-Pro global accuracy.

**FIGURE 22**
EMA routing MMLU-Pro latency.

TABLE 7 Comparison of EMA in MMLU and MMLU-Pro.

| Benchmark | Model setting | Accuracy | Avg latency |
|---|---|---|---|
| MMLU | ORCH | 0.813 | 11,775 |
| MMLU | ORCH + EMA-router | 0.820 | 12,315 |
| MMLU-Pro | ORCH | 0.733 | 18,263.4 |
| MMLU-Pro | ORCH + EMA-router | 0.753 | 19,645.7 |

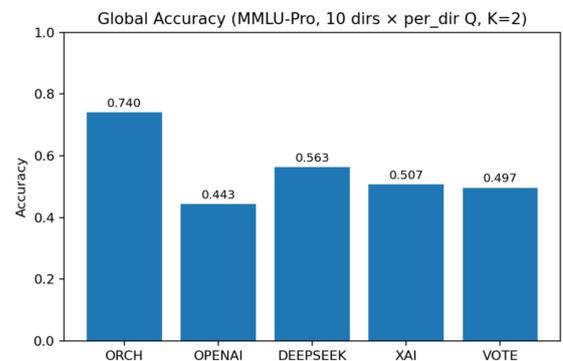

**FIGURE 23**
Self-consistency MMLU-Pro global accuracy.

accuracy relative to ORCH + EMA and increases latency by more than a factor of four, making this configuration more suitable as a high-budget upper bound than as a default deployment option.

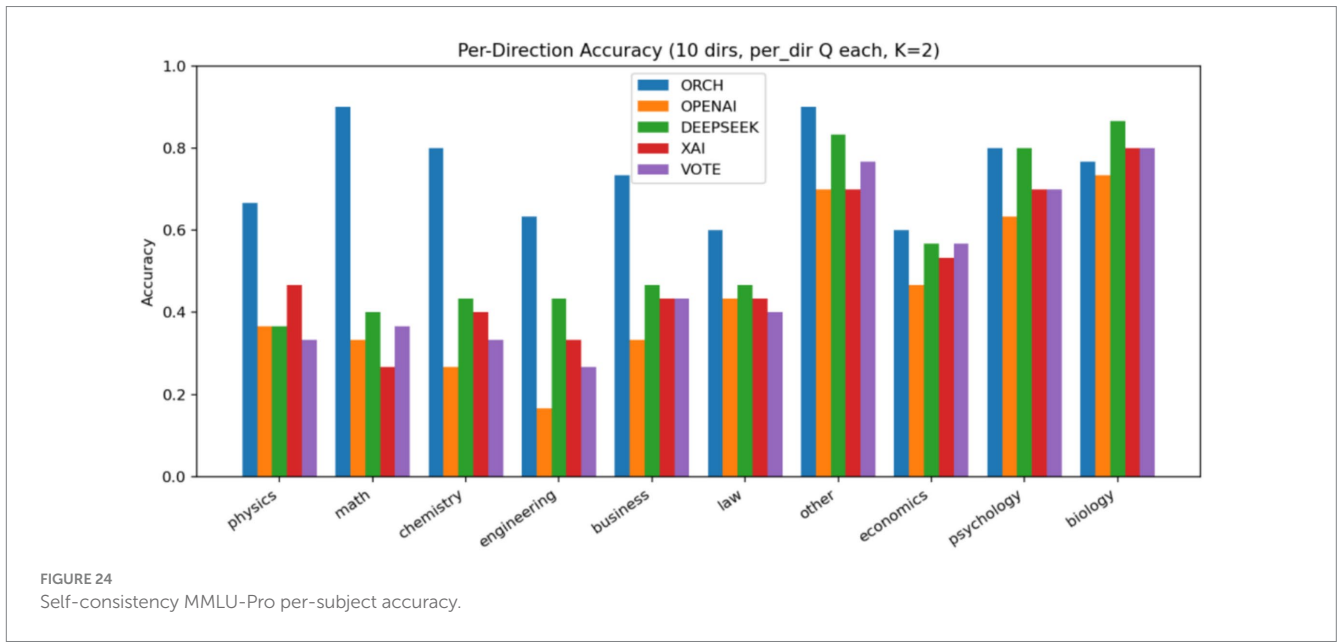

FIGURE 24
Self-consistency MMLU-Pro per-subject accuracy.

FIGURE 25
Self-consistency MMLU-Pro latency.

TABLE 8 Self-consistency comparison table.

| Benchmark | Model setting | Accuracy | Avg latency (ms/question) |
|---|---|---|---|
| MMLU-Pro | ORCH + EMA-router | 0.753 | 19,645.7 |
| MMLU-Pro | ORCH + EMA-router + self-consistency ($K = 2$) | 0.740 | 89,462.0 |

From a cost perspective, if O, D, and X denote the per-call token costs of the three base models, the cost of VOTE is approximately O + D + X. Three-agent ORCH, which requires "multi-model analysis + a merging agent for second-stage reasoning," is roughly $3 \times (O + D + X)$; with the consistency module enabled, the effective cost can be regarded as on the order of $6 \times (O + D + X)$. In practical applications, one can therefore choose along a spectrum from "single model → VOTE → ORCH-1 → ORCH (+EMA) → ORCH (with consistency)" according to the desired balance between accuracy and latency/cost. A consolidated comparison of these configurations is reported in Table 9.

# 5 Discussion and summary

## 5.1 Summary

Guided by the "many analyses, one merge" paradigm, this study proposed ORCH, a protocol-level deterministic multi-agent coordination framework tailored to discrete-choice reasoning. Without retraining any underlying large language model, ORCH relies solely on a pipeline of "parallel multi-model analysis + a single merging agent for final decision-making" and was systematically evaluated on three benchmarks: MMLU, MMLU-Pro, and GSM8K. Across these settings—covering standard four-option multiple choice, more demanding ten-option multi-domain questions, and multi-step numerical reasoning—ORCH consistently and substantially outperformed three strong single-model baselines (OpenAI gpt-4o-mini, DeepSeek-chat, XAI Grok-2-latest) and a simple majority-vote scheme (VOTE). Although the MMLU results do not pass the statistical significance test, the point estimates still show a small upward shift. The gains are especially pronounced on MMLU-Pro and GSM8K, where ORCH exceeds the strongest single model by several tens of percentage points; McNemar tests confirm that these improvements are statistically significant, supporting the effectiveness of protocol-level deterministic multi-agent coordination in complex reasoning scenarios.

A series of comparative and ablation experiments further clarify the sources of ORCH performance and the associated cost–benefit profile. Removing one agent (XAI) to form ORCH-1 leads to a noticeable drop in overall accuracy on MMLU-Pro, indicating that diversity and complementarity among multiple agents are key contributors to the performance gains, rather than reliance on a single strongest baseline. Although introducing EMA does not yield statistical significance, it is still associated with a small increase in the point estimate. Introducing EMA-based routing allows ORCH to dynamically choose which base model acts as the decomposition layer and which as the merger, based on historical correctness, latency, and cost. This yields a small but consistent accuracy gain of about 0.7–2.0 percentage points on MMLU and MMLU-Pro, with 5–10% additional latency and token cost; the improvement over the best single-model baseline is statistically significant in McNemar tests, whereas the paired comparison with fixed-routing ORCH shows a more modest "consistently better" trend. By contrast, augmenting ORCH + EMA with self-consistency and multi-shuffle at the merge layer retains a clear advantage over single models and VOTE, but does not further increase global accuracy under the current settings and instead inflates average latency by several times. This suggests that a "strong consistency + multi-shuffle" configuration is better interpreted as a high-budget upper bound rather than a practical default strategy.

TABLE 9 All experiments compared.

| Benchmark | ORCH variants/configurations | Accuracy | Avg latency (ms/question) | Accuracy vs strongest single model (percentage points) |
| --- | --- | --- | --- | --- |
| MMLU | ORCH (3 agents, fixed route) | 0.813 | 11,775 | +4.6 |
| MMLU | ORCH + EMA-router | 0.820 | 12,315 | +7.0 |
| MMLU-Pro | ORCH (3 agents, fixed route) | 0.733 | 18,263.4 | +17.6 |
| MMLU-Pro | ORCH-1 (No XAI, OpenAI + DeepSeek) | 0.700 | 16,873.8 | +14.0 |
| MMLU-Pro | ORCH + EMA-router | 0.753 | 19,645.7 | +19.0 |
| MMLU-Pro | ORCH + EMA routing + consistency ($K = 2$, $m = 1$) | 0.740 | 89,462.0 | +17.7 |
| GSM8K | ORCH (3 agents, fixed route) | 0.940 | 11,453.5 | +50.0 |

Despite the favourable results of ORCH and the EMA router on multiple benchmarks, several limitations remain. First, the present evaluation focuses on choice-style reasoning benchmarks and does not yet cover richer task types such as code generation, open-ended QA, or tool-augmented interaction. Second, the framework currently integrates only three mainstream commercial models; it does not systematically explore how the number of models, capability gaps, or instruction styles influence orchestration quality, and larger, more heterogeneous model pools should be examined in future work. Third, EMA routing assumes an offline, labelled evaluation setting and thus relies on explicit correctness feedback; its design is not yet adapted to real-world "unlabelled or weak-feedback" environments.

Taken together, our findings suggest that even as large models approach a "platform-level" performance ceiling, careful protocol-level deterministic orchestration of multiple agents, combined with dynamic routing and appropriate statistical analysis, can still unlock meaningful gains in reasoning quality without additional training. This points to a viable path toward building controllable, interpretable, and practically deployable agentic systems on top of existing LLM infrastructure.

## 5.2 Limitations

ORCH has several limitations that need to be recognized. First, because the system relies on external large language model (LLM) APIs, it is subject to operational restrictions such as rate limits, temporary outages, changes in model versions, and fluctuations in cost and latency. These factors can impact throughput and consistency during extended evaluations. However, these issues can be addressed by using techniques like caching prompt–response pairs, implementing controlled protocol-level deterministic retries, and having explicit fallback modes (for example, switching from multi-agent to a single default agent when the system is congested), along with logging model identifiers and configuration details. Second, coordinating multiple agents does not eliminate bias and may even increase shared biases if the models involved have similar training data or safety guidelines. Additionally, protocol-level deterministic merging methods (such as confidence-based tie-breakers) can unintentionally favor certain models or response styles. To counter this, it is important to monitor bias systematically through stratified evaluations and disagreement analyses, conduct regular audits of merge results, and diversify the set of models used when possible. Third, although ORCH shows improved reliability on discrete-choice benchmarks, it should not replace expert judgment in safety-critical or regulated areas because protocol-level deterministic aggregation can still produce confidently incorrect answers when ambiguity exists. Therefore, high-risk applications should include human oversight triggered by specific risk indicators (like low confidence, significant disagreement among agents, or schema violations) and keep comprehensive logs of subquestions, agent responses, and merging decisions to enable auditing and ensure accountability.

Because ORCH relies on multiple external LLM APIs, the measured latency and token-based cost can fluctuate due to network conditions, transient provider load, and request-level variations. In addition, token accounting is more complex than a simple text-length proxy: tokenization differs by model, usage metadata may be reported at different granularities (prompt vs. completion vs. total), and some systems include additional internal "reasoning" tokens. For these reasons, the present study reports point estimates and a limited variability analysis under a budgeted evaluation setting. In future work, we will (i) run repeated trials on a fixed subset to report mean and standard deviation (or confidence intervals) for latency and token-cost, (ii) log usage metadata from API responses in a standardized way, and (iii) document rate-limit and provider-version factors that may affect reproducibility.

## 6 Future work

Future research may proceed along several directions. (i) Extending ORCH to a broader range of tasks and languages,

including code reasoning, retrieval-augmented QA, and multi-turn dialogue, to test its generality. (ii) Evolving the current heuristic EMA routing into a learnable or reinforcement-learning-based policy, augmented with uncertainty estimates and confidence calibration, so that routing decisions become more fine-grained and adaptive. (iii) Distilling the high-quality reasoning traces produced by multi-agent orchestration into a single student model, aiming for an "offline multi-agent, online single-model" deployment pattern that reduces runtime cost. (iv) Incorporating richer evaluation dimensions—such as human quality ratings, robustness and safety metrics—to more comprehensively characterise the benefits and risks of multi-agent coordination in realistic settings.

In fields like healthcare, finance, and public services, system failures often result from low-frequency, high-impact events rather than average-case performance. These events can include adversarial manipulation and gaps in traceability. The risks increase in situations involving agents or multiple agents, as using tools and executing multiple steps can enlarge the attack surface. Examples of this are prompt injection, instruction hijacking, and unsafe action execution. Recent studies of LLM-based agents show that these agents might comply with misuse requests. They can also be vulnerable to injected instructions and can be directed towards actions chosen by attackers in controlled environments.

A key focus for future work is to strengthen safety measures around ORCH-style orchestration in high-stakes situations. This includes: (i) input-side constraints like schema validation, policy filtering, and grounding retrieval with consistency checks; (ii) output-side verification, which involves cross-checking, detecting contradictions, and calibrated refusal under high uncertainty; (iii) human oversight for important decisions; and (iv) auditable records that track routing choices, trigger events, and clear output summaries in a minimally sensitive format. This will help support accountability after the fact, possibly using tamper-evident logging or anchoring. These steps follow established risk-management advice, which highlights the importance of governance, documentation, monitoring, and preventing misuse of generative AI in safety-critical areas.

For high-stakes settings such as healthcare, finance, and public administration, trustworthy operation often depends less on average performance and more on auditability, accountability, and tamper-evident process evidence. Inspired by the landscape of AI for blockchain summarized by Ressi et al. (2024) future work may treat blockchain as a trusted audit substrate for securing the model routing and orchestration workflow. Without disclosing sensitive content, key decisions, including route selection, threshold triggers, and output digests, can be committed on chain as hashed traces, where timestamps and immutability provide end-to-end evidence preservation. A further direction is to explore AI-enhanced on-chain auditing policies, for example, using models to determine which events warrant finer-grained tracking and to escalate audit levels when anomalous patterns are detected, thereby balancing audit overhead against governance value. This line of work primarily targets governance and trust: it provides directionally stronger verifiability rather than a claim of universal advantage across configurations and deployment contexts.

## Data availability statement

The original contributions presented in the study are included in the article/Supplementary material, further inquiries can be directed to the corresponding author.

## Author contributions

HZ: Methodology, Supervision, Writing – original draft, Writing – review & editing. HC: Validation, Writing – review & editing.

## Funding

The author(s) declared that financial support was not received for this work and/or its publication.

## Conflict of interest

The author(s) declared that this work was conducted in the absence of any commercial or financial relationships that could be construed as a potential conflict of interest.

## Generative AI statement

The author(s) declared that Generative AI was not used in the creation of this manuscript.

Any alternative text (alt text) provided alongside figures in this article has been generated by Frontiers with the support of artificial intelligence and reasonable efforts have been made to ensure accuracy, including review by the authors wherever possible. If you identify any issues, please contact us.

## Publisher's note

All claims expressed in this article are solely those of the authors and do not necessarily represent those of their affiliated organizations, or those of the publisher, the editors and the reviewers. Any product that may be evaluated in this article, or claim that may be made by its manufacturer, is not guaranteed or endorsed by the publisher.

## Supplementary material

The Supplementary material for this article can be found online at: https://www.frontiersin.org/articles/10.3389/frai.2026.1748735/full#supplementary-material